\documentclass[11pt]{article}

\usepackage{fullpage}
\usepackage{xfrac}

\usepackage[]{color-edits}
\usepackage{amsmath,graphicx}

\usepackage[utf8]{inputenc} 
\usepackage[T1]{fontenc}    
\usepackage{hyperref}       
\usepackage{url}            
\usepackage{booktabs}       
\usepackage{amsfonts}       
\usepackage{nicefrac}       
\usepackage{microtype}      

\usepackage{color}
\usepackage{subfigure} 
\usepackage{graphicx}
\usepackage{wrapfig}

\usepackage{multirow}
\usepackage{makecell}
\usepackage{threeparttable}

\usepackage[linesnumbered,algoruled,boxed,lined,algo2e]{algorithm2e}
\usepackage{amsthm}
\usepackage{amsmath}
\usepackage{amssymb}

\newtheorem{assume}{Assumption}

\newtheorem{theorem}{Theorem}
\newtheorem{lemma}{Lemma}
\newtheorem*{theorem-non}{Theorem}

\title{Learning Fast Approximations of Sparse Nonlinear Regression}

\newcommand*{\email}[1]{\texttt{#1}}
\author{
    Yuhai Song$^{1}$,
    Zhong Cao$^{1}$, 
    Kailun Wu,
    Ziang Yan$^{1}$ and
    Changshui Zhang$^{1}$}
\date{
    $^{1}$ Institute for Artificial Intelligence, Tsinghua University (THUAI)\\
    Beijing National Research Center for Information Science and Technology (BNRist)\\
    Department of Automation,Tsinghua University, Beijing, P.R.China \\
\email{
    \{song-yh19, caozhong14, yza18\}@mails.tsinghua.edu.cn\\
    whywkl305@qq.com \quad zcs@mail.tsinghua.edu.cn
    }
}

\begin{document}
%
\maketitle
\begin{abstract}
The idea of unfolding iterative algorithms as deep neural networks has been widely applied in solving sparse coding problems, providing both solid theoretical analysis in convergence rate and superior empirical performance.
 However, for sparse nonlinear regression problems, a similar idea is rarely exploited due to the complexity of nonlinearity.
 In this work, we bridge this gap by introducing the \emph{Nonlinear Learned Iterative Shrinkage Thresholding Algorithm} (NLISTA), which can attain a linear convergence under suitable conditions.
 Experiments on synthetic data corroborate our theoretical results and show our method outperforms state-of-the-art methods.
 The source code is available at https://github.com/songyh15/NonlinearLISTA.
\end{abstract}

\section{Introduction}

In this paper, we aim to estimate a sparse vector ${x}^* \in \mathbb{R}^{n}$ from its noisy nonlinear measurement ${y} \in \mathbb{R}^m$:
\begin{equation}
	\label{nonlinear system} 
	y = f(Ax^*) + \varepsilon,
\end{equation}
where $A \in \mathbb{R}^{m \times n}$, $m \ll n$, $\varepsilon \in \mathbb{R}^m$ is the exogenous noise, and $f(\cdot)$ is an element-wise nonlinear function.
As directly minimize the $\ell_0$-norm to promote the sparsity is shown to be NP-hard~\cite{blumensath2008iterative}, the sparsity is typically achieved via minimizing the least square error augmented by $\ell_{1}$-regularization instead.

For solving sparse nonlinear regression problems, the SpaRSA (sparse reconstruction by separable approximation) method~\cite{wright2009sparse}
minimizes the upper bound of the $\ell_{1}$-regularized objective iteratively
by using a simple diagonal Hessian approximation,
which results in an iterative shrinkage thresholding algorithm.
The fast iterative soft thresholding algorithm (FISTA)~\cite{beck2009fast}
accelerates the convergence of iterations by using a very specific linear combination of the previous two outputs as the input of the next iteration.
The fixed point continuation algorithm (FPCA)~\cite{hale2008fixed}
lowers the shrinkage threshold value in a continuation strategy,
which makes the iterative shrinkage thresholding algorithm converge faster.
The iterative soft thresholding with a line search algorithm (STELA)~\cite{yang2018parallel}
uses a line search scheme to calculate the step size for updating the input of the next iteration.
Despite the fact that the $\ell_1$-regularized objective is nonconvex in general due to the nonlinearity of $f(\cdot)$,
\cite{yang2016sparse} proved that under mild conditions, with high probability the stationary points of the SpaRSA method are close to the global optimal solution.

Over the last decade, the community has made massive efforts in developing deep unfolding methods to solve sparse regression problems efficiently in a special case where $f(\cdot)$ is the identity function, in which~\eqref{nonlinear system} is reduced to the well known sparse coding model.
An early attempt named Learned Iterative Shrinkage Thresholding Algorithm (LISTA)~\cite{gregor2010learning} proposed to unfold the Iterative Shrinkage-Thresholding Algorithm (ISTA) using deep neural networks with learnable weights, whose requisite numbers of iterations to obtain similar estimation results are one or two orders of magnitude less than that of ISTA.
There are also another kind of learning-based ISTA called ISTA-Net \cite{zhang2018ista}
and some improved versions of LISTA, such as 
TISTA \cite{ito2019trainable},
Step-LISTA \cite{ablin2019learning}, LISTA-AT \cite{kim2020element},
and GLISTA \cite{wusparse}.
Although the deep unfolding technique is promising \cite{hershey2014deep}, 
there is no learning-based approach that can deal with nonlinear cases due to the complexity caused by nonlinearity.
As a more generalized case of~\eqref{nonlinear system}, the sparse recovery problem over nonlinear dictionaries also gains some attention~\cite{chamon2019sparse,chamon2020functional}.
However, those deep unfolding methods are not directly applicable to nonlinear dictionaries due to the different formulation.

In this paper, we aim to exploit the idea of unfolding classical iterative algorithms as deep neural networks with learnable weights to solve the sparse nonlinear regression problem.
To the best of our knowledge,
our proposed Nonlinear Learned Iterative Shrinkage Thresholding Algorithm (NLISTA) is the first deep sparse learning network for the sparse nonlinear regression problem.
We provide theoretical analysis to show that under mild conditions, there exists a set of parameters for the deep neural network that could ensure NLISTA converges at a linear rate to the ground truth solution.
Experimental results on synthetic data corroborate our theoretical results and show our method outperforms state-of-the-art sparse nonlinear regression algorithms.

\section{Algorithm Description}

\begin{figure}[t]
    \centering
    \includegraphics[width=1\linewidth]{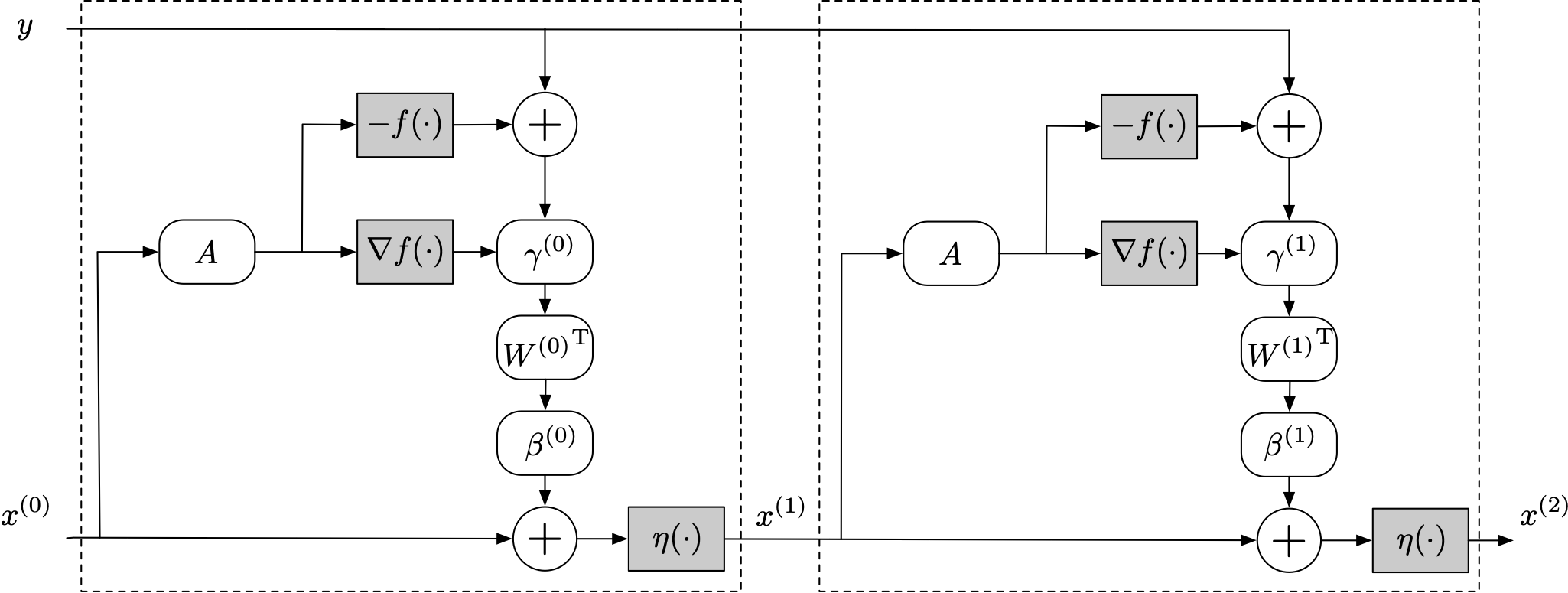}
    \caption{Network architecture of NLISTA with T = 2.}
    \label{fig:NLISTA}
\end{figure}

The iterative step of the SpaRSA method \cite{yang2016sparse} \cite{wright2009sparse}
can be formulated as:
\begin{equation}
    \label{SpaRSA}
    x^{(t+1)} = \eta(x^{(t)} - \frac{1}{\alpha^{(t)}} \nabla L(x^{(t)}), \frac{\lambda}{\alpha^{(t)}} )
\end{equation}
where $t$ represents the $t$-th iteration,
$ \eta(u, a) := sign(u) max$
$\{|u|-a, 0\}$ is the soft thresholding function,
$L(x) := \frac{1}{2}\|y-f(Ax)\|_{2}^{2}$ is the least square loss function,
$\lambda$ is a scalar representing the $\ell_{1}$-regularization parameter
and $\alpha^{(t)}$ is a constant larger than the largest eigenvalue of $\nabla^2 L(x^{(t)})$.
$\nabla$ represents the gradient and $\nabla^2$ represents the Hessian matrix.
Based on the relationship between $\nabla L(x^{(t)})$ and $\nabla f(Ax^{(t)})$,
\begin{equation}
    \label{grad_L and grad_f}
    \nabla L(x^{(t)}) = A^\mathrm{T} \nabla f(Ax^{(t)})(f(Ax^{(t)}) - y),
\end{equation}
we convert (\ref{SpaRSA}) to:
\begin{equation}
    \label{SpaRSA_f}
    x^{(t+1)} = \eta(x^{(t)} + \frac{1}{\alpha^{(t)}} A^\mathrm{T} \nabla f(Ax^{(t)})(y - f(Ax^{(t)})), \frac{\lambda}{\alpha^{(t)}} )
\end{equation}
Furthermore, we propose the Nonlinear Learned Iterative Shrinkage Thresholding Algorithm (NLISTA),
whose iterative step can be formulated as:
\begin{equation}
    \label{NLISTA}
        x^{(t+1)} = \eta(x^{(t)} + \beta^{(t)} {W^{(t)}}^\mathrm{T} \gamma^{(t)} \nabla f(Ax^{(t)})(y - f(Ax^{(t)})), \theta^{(t)})
\end{equation}
where $W^{(t)} \in \mathbb{R}^{m \times n}$, $\beta^{(t)} \in \mathbb{R}$ and $\theta^{(t)} \in \mathbb{R}$
are free parameters to train,
and $\gamma^{(t)} \in \mathbb{R}$ is defined as:
\begin{equation}
    \label{gamma}
    \gamma^{(t)} = \left\{
        \begin{array}{ll}
            1 \quad \quad \quad ,\|\nabla f(Ax^{(t)})(y - f(Ax^{(t)})) \|_{2} \le 1\\
            \|\nabla f(Ax^{(t)})(y - f(Ax^{(t)})) \|_{2}^{-1} ,otherwise.\\
        \end{array}
        \right.
\end{equation}
The network architecture of NLISTA is illustrated in Fig.\ref{fig:NLISTA},
which remains the recurrent neural network structure.

The effect of $\gamma^{(t)}$,
which is not trainable,
is to restrict the product of the gradient item $\nabla f(Ax^{(t)})$
and the residual item $y - f(Ax^{(t)})$ when the product is too large.
The item $(W^{(t)})^\mathrm{T} \gamma^{(t)} \nabla f(Ax^{(t)})(y - f(Ax^{(t)}))$ 
represents the updating direction,
which is supposed to be close to zero when the $\ell_{2}$-norm of the residual item is small enough.
So the product of the gradient item and the residual item will not be normalized 
if the norm of the product is small enough,
where we let $\gamma^{(t)}$ equal one.

The effect of $W^{(t)}$ can be viewed as adjusting the updating direction based on the training data.
The item $\beta^{(t)}$ can be treated as the updating step size,
whose effect is similar to the item $\frac{1}{\alpha^{(t)}}$ in the SpaRSA method.
Since the SpaRSA method may converge too slow if $\alpha^{(t)}$ is too large
and may not be able to converge to the global optimal solution if $\alpha^{(t)}$ is too small,
the step size is set to be trainable in NLISTA considering its great influence on the convergence property.



As the regularization parameter, $\lambda$ has a significant impact on the sparsity degree of the results.
Due to the effect of the shrinkage thresholding function,
the output of the SpaRSA method will be sparser when $\lambda$ is set larger.
However, it is hard to set a proper regularization parameter to attain a certain sparsity degree of the outputs
which is supposed to be consistent with the data.
In NLISTA, we let the threshold value $\frac{\lambda}{\alpha^{(t)}}$ be trainable and denote it as $\theta^{(t)}$,
which means that the regularization parameter, as well as the sparsity degree of the outputs, 
is determined by the data instead of human intervention.
As a result, a more suitable threshold vale of each iteration will be chosen in NLISTA.

Experiment results in Fig.\ref{fig:cosx} illustrate that
the performance of NLISTA is much better than existing state-of-the-art algorithms.
Concretely, NLISTA does not only converge faster but also has a much smaller recovery error.

\section{Convergence Analysis}

In this section, we analyze the convergence property of NLISTA. 
We first state two following assumptions on samples and dictionary matrices before presenting the main theorem.

\begin{assume}
    \label{assume:x & epi}
    The sparse vector $x^*$ and  the exogenous noise $\varepsilon$ belong to the following sets:
      \begin{equation}
        \label{eq:x_assume}
        \Omega_{x}(c_x,s) \triangleq  \Big\{ x^* \in \mathbb{R}^{n} \Big| \|x^\ast \|_{\infty} \leq c_x, \|x^\ast\|_0 \leq s \Big\},
      \end{equation}
      \begin{equation}
        \Omega_{\varepsilon}(\sigma) \triangleq  \Big\{ \varepsilon \in \mathbb{R}^{m} \Big| \| \varepsilon \|_1 \le \sigma  \Big\},
      \end{equation}
      where $c_x > 0$,$s>0$ and $\sigma > 0 $ are constants.
\end{assume}
Therefore, there are upper bounds for each value of $x^*$, the amount of the non-zero elements, 
and the $\ell_{1}$-norm of $\varepsilon$,
which are actually common conditions.

\begin{assume}
  \label{assume:A}
  The dictionary matrix $A$ belongs to the following set:
    \begin{equation}
      \label{eq:A_assume}
      \begin{array}{ll}
      \Omega_{A} \triangleq  \Big\{ A \in \mathbb{R}^{m \times n} \Big| 
      & A_i^\mathrm{T} A_i = 1, \mathop{\rm max}\limits_{i \neq j} | A_i^\mathrm{T} A_j | < 1,\\
      & i,j= 1,2,\cdots,n \Big\},
      \end{array}
    \end{equation}
    where $A_i$ represents the $i^{th}$ column of $A$.
\end{assume}
Therefore, the dictionary matrix $A$ is required to be column normalized and constrained in the column correlation.
Then, we need to introduce the following lemmas as the preparation for the main theorem.
\begin{lemma}
  \label{lemma: mean value theorem}
  $Lagrange$ $mean$ $value$ $theorem.$
  \quad
  If $f(\cdot)$ is continuously differentiable in $[-c_x,c_x]$, then for any t,
  there exists $\xi^{(t)} \in \mathbb{R}^{m}$ subject to
  \begin{equation}
      f(Ax^*) - f(Ax^{(t)}) = \nabla f(\xi^{(t)})(Ax^* - Ax^{(t)}).
    \label{lemma_1}
  \end{equation}
\end{lemma}
Since the nonlinear function $f(\cdot)$ is element-wise,
the Lagrange mean value theorem will holds when $f(\cdot)$ is continuously differentiable in $[-c_x,c_x]$.
For expression simplicity, hereinafter all $\xi^t$ used in equations satisfy (\ref{lemma_1}).

\begin{lemma}
  \label{lemma: W}
  If $f(\cdot)$ is continuously differentiable in $[-c_x,c_x]$, 
  the gradient of $f(\cdot)$ is nonzero for any $x \in [-c_x,c_x]$,
  Assumption \ref{assume:x & epi} holds and Assumption \ref{assume:A} holds,
  then $\Omega_W^{(t)}$ is not an empty set,
  where 
  \begin{equation}
    \begin{aligned}
    &\Omega_W^{(t)} \triangleq \Big\{ W \in \mathbb{R}^{m \times n} \Big|
          \beta^{(t)} \gamma^{(t)} W_i^\mathrm{T} \nabla f(Ax^{(t)}) \nabla f(\xi^{(t)}) A_i = 1,
          \\
          &\mathop{\rm max}\limits_{i \neq j} 
         | \beta^{(t)} \gamma^{(t)} W_i^\mathrm{T} \nabla f(Ax^{(t)}) \nabla f(\xi^{(t)}) A_j | < 1,
         i,j= 1,\cdots,n \Big\}
    \end{aligned}
  \end{equation}
\end{lemma}
The proof of Lemma \ref{lemma: W} can be found in the supplementary.
Lemma \ref{lemma: W} actually describes a specific matrix set which is critical for the following theorem.

\begin{lemma}
  \label{lemma: no false positive}
  If $f(\cdot)$ is continuously differentiable in $[-c_x,c_x]$, 
  $x^{(0)} = 0$, $\{x^{(t)}\}_{t=1}^{\infty}$ are generated by (\ref{NLISTA}),
  Assumption \ref{assume:x & epi} holds,
  and $\theta^{(t)} \ge \mu_1^{(t)} \|x^* - x^{(t)} \|_1 + \mu_2^{(t)} \sigma$ ,
  then
  \begin{equation}
      x_i^{(t)} = 0, \quad \forall i \notin S, \quad \forall t \ge 0,
  \end{equation}
  where 
  \begin{equation}
      \begin{array}{ll}
      \mu_1^{(t)} = \mathop{\rm max}\limits_{i \neq j} | \beta^{(t)} \gamma^{(t)} W_i^\mathrm{T} \nabla f(Ax^{(t)}) \nabla f(\xi^{(t)}) A_j | ,
      \\
      \mu_2^{(t)} = \mathop{\rm max}\limits_{i} \| \beta^{(t)} \gamma^{(t)} W_i^\mathrm{T} \nabla f(Ax^{(t)}) \|_1 ,
      i,j= 1,2,\cdots,n,
      \end{array}
  \end{equation}
  and $S \triangleq \Big\{ i \Big| x_i^*  \neq 0  \Big\}$ is the support set of $x^*$.
\end{lemma}
The proof of Lemma \ref{lemma: no false positive} can be found in the supplementary.
Lemma \ref{lemma: no false positive} actually describes a simple fact that
some elements of each iteration output will keep zero as long as the shrinking threshold is large enough.
And the specific constants defined in Lemma \ref{lemma: no false positive}, 
$\mu_1^{(t)}$ and $\mu_2^{(t)}$,
will be used in the following theorem 
and are critical for the theorem proof.
Contrasting the constants definitions and the matrix set constraints in Lemma \ref{lemma: W},
we can find that the constants are some kind of evaluation of the learned matrix $W$,
which are expected to be as small as possible.

We now are ready to introduce following main theorem about the convergence property of  NLISTA.
\begin{theorem}
    \label{theorem:1}
    If $f(\cdot)$ is continuously differentiable in $[-c_x,c_x]$,
    $x^{(0)}=0$, 
    Assumption \ref{assume:x & epi} holds, Assumption \ref{assume:A} holds
    and $\{x^{(t)}\}_{t=1}^{\infty}$ are generated by (\ref{NLISTA}),
    then there exists a set of parameters $\{W^{(t)}, \theta^{(t)}\}_{t=0}^{\infty}$ where
    $W^{(t)} \in \Omega_W^{(t)}$ and
    $\theta^{(t)} \ge \mu_1^{(t)} \|x^* - x^{(t)} \|_1 + \mu_2^{(t)} \sigma$ for any $t \ge 0$,
    such that
    \begin{equation}
        \label{eq:linear_conv}
        \|x^{(t)}-x^\ast\|_2 \leq s c_x q^{t} + 2s c_{\varepsilon}\sigma,\quad \forall t = 0,1,\cdots,
    \end{equation}
    where $q$ and $c_{\varepsilon}$ are constants that depend on 
    $\{\mu_1^{(t)}\}_{t=0}^{T}$, $\{\mu_2^{(t)}\}_{t=0}^{T}$ and $s$.
    $q \in (0,1)$ if $s$ is sufficiently small, and $c_{\varepsilon} > 0$.
    The definitions are omitted due to space limitations and can be found in the arXiv version of the paper.
    Note that the $t$ in $q^t$ is the exponent.
\end{theorem}

If $\sigma=0$, (\ref{eq:linear_conv}) reduces to
\begin{equation}
    \label{eq:linear_noiseless}
    \|x^{(t)}-x^\ast\|_2 \leq s c_x q^t, \quad \forall t = 0,1,\cdots.
\end{equation}

The proof of Theorem \ref{theorem:1} can be found in the supplementary.
Theorem \ref{theorem:1} means that in the noiseless case,
there exist parameters enabling the upper bound of the NLISTA estimation error to converge to zero
at a $q$-linear rate with the number of layers going to infinity.
As a result of the convergence property of the upper bound, the NLISTA estimation error also converges to zero quickly,
which is validated by Figure \ref{fig:cosx_noiseless}.

Theorem \ref{theorem:1} also demonstrates that
the existence of the noise will increase the upper bound of the NLISTA estimating error.
And the convergence speed of NLISTA under noisy conditions is also linear,
which is illustrated in Figure \ref{fig:cosx_SNR30}.

Due to the relationship between $q$ and the upper bound of $ |\nabla f(\cdot) |$,
we can derive that the upper bound of the estimating error will converge slower when the upper bound of $|\nabla f(\cdot)|$ is larger
based on Theorem \ref{theorem:1}.
As a result, the performance of NLISTA is supposed to be better for the nonlinear function $f(\cdot)$
whose supremum of $|\nabla f(\cdot)|$ is smaller,
which is validated by Table \ref{table: gradient}.

\section{Experiments}

\begin{figure*}[t]
  \centering
  \vspace{-1em}
  \begin{tabular}{ccc}
    \hspace{-1.5em}
    \subfigure[Noiseless Case: $\textrm{SNR}$=$\infty$]{
      \includegraphics[width=0.33\linewidth]{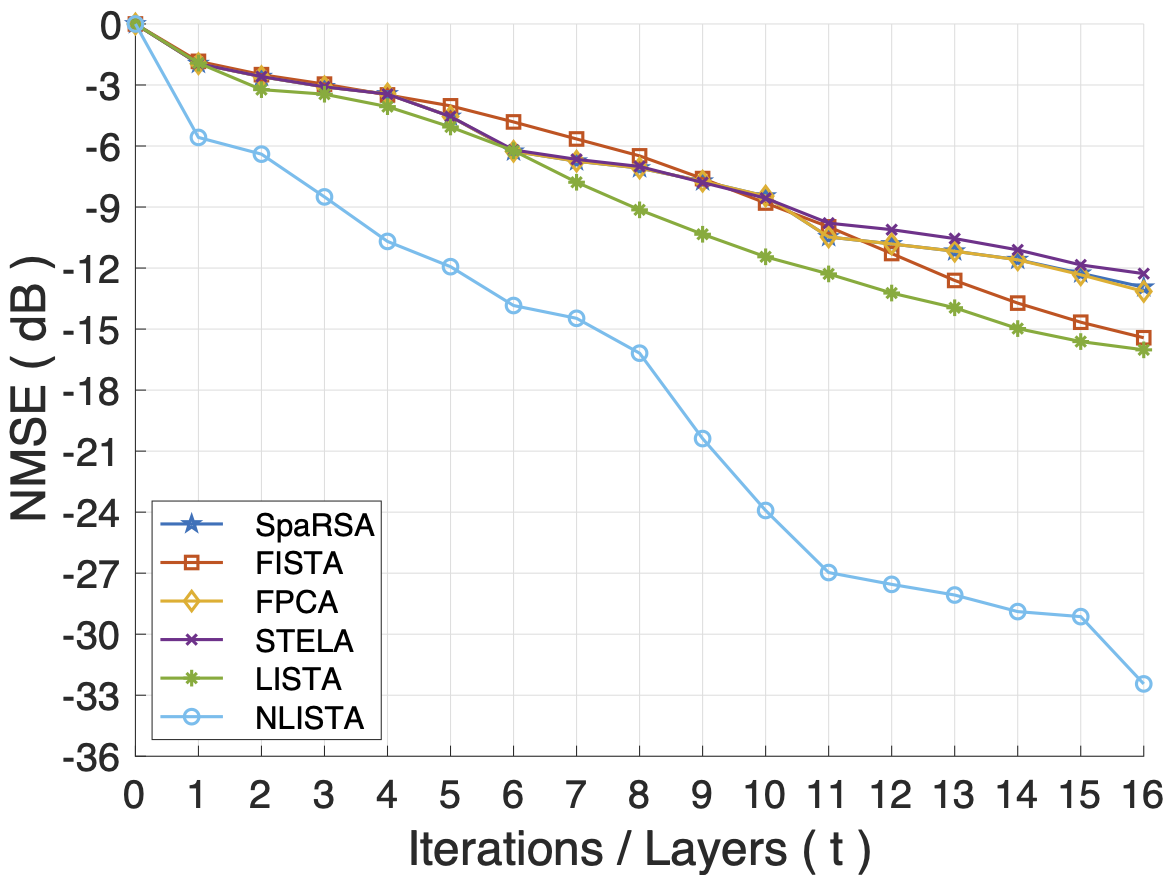}
      \label{fig:cosx_noiseless}
    }
    &
    \hspace{-1.5em}
    \subfigure[Noisy Case: $\textrm{SNR}$=30dB]{
      \includegraphics[width=0.33\linewidth]{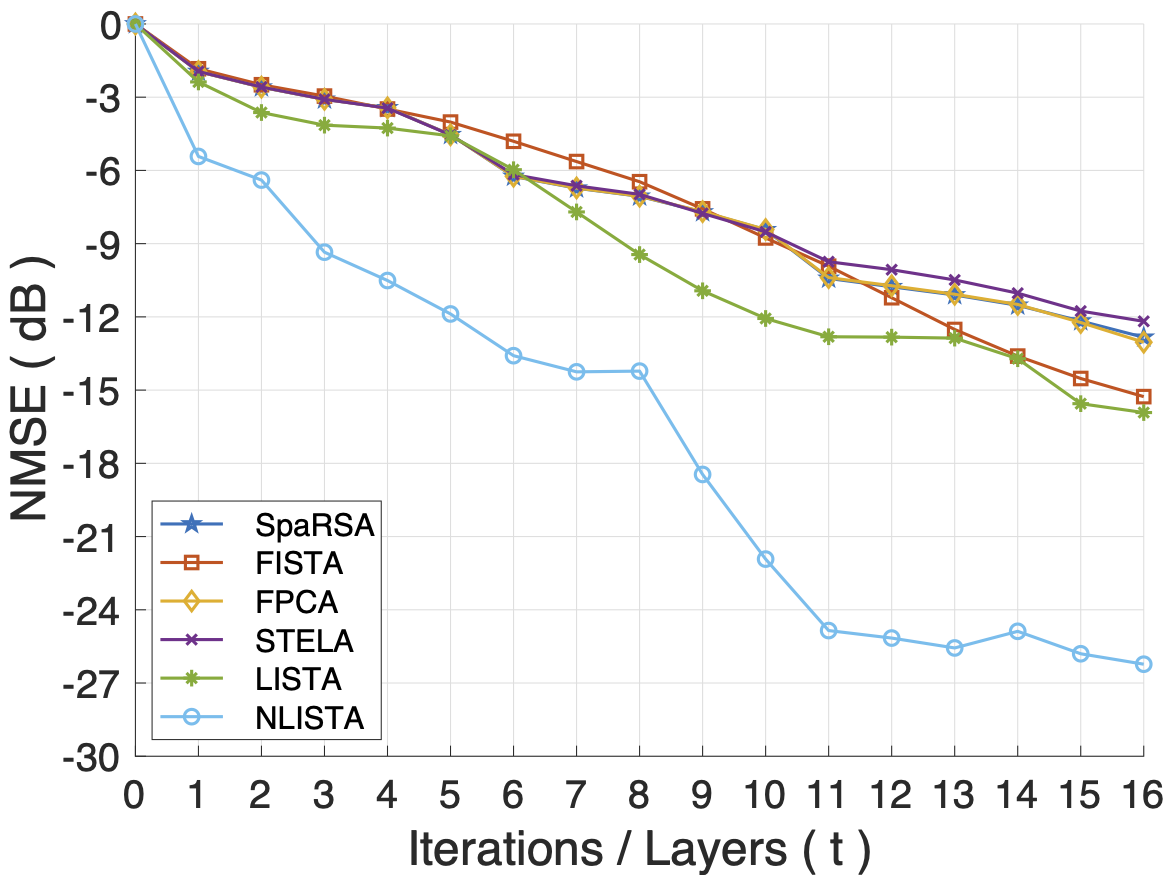}
      \label{fig:cosx_SNR30}
    }
    &
    \hspace{-1.5em}
    \subfigure[Performance with ill-conditioned matrix]{
      \includegraphics[width=0.33\linewidth]{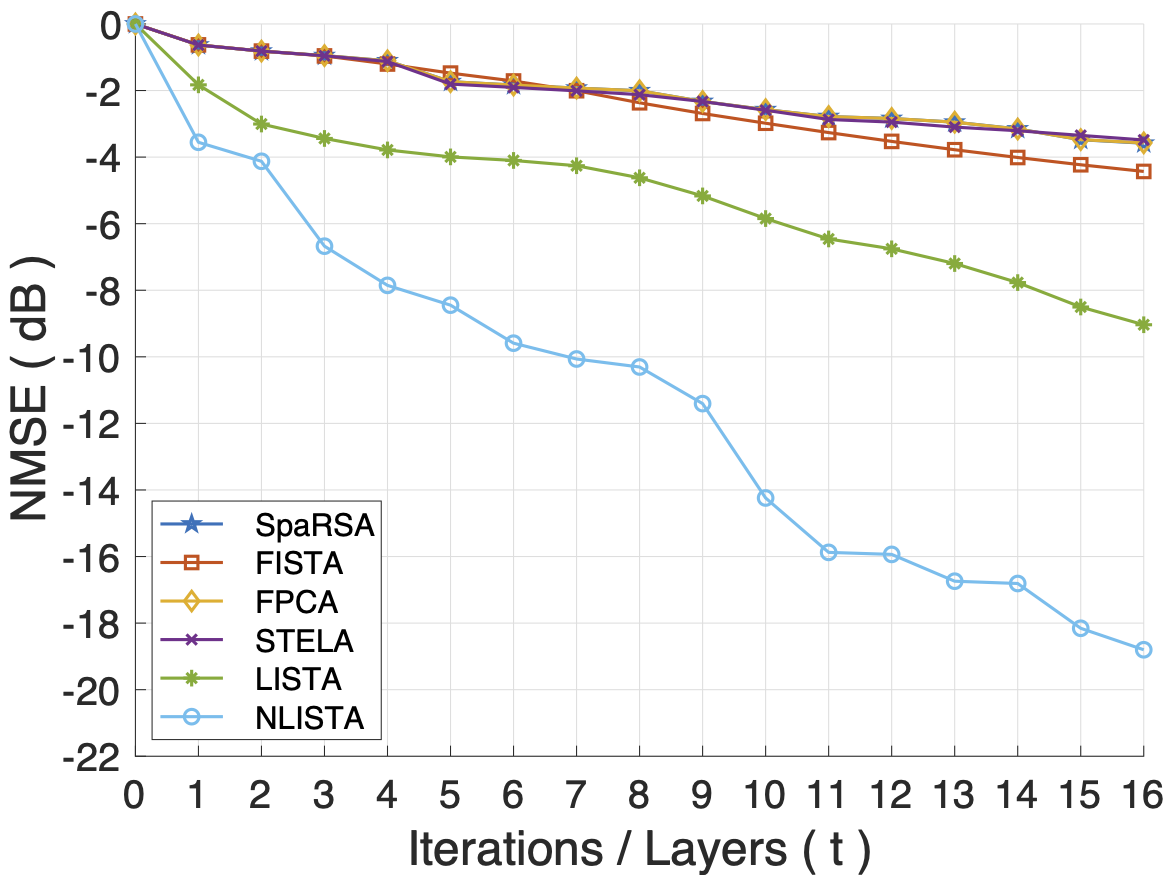}
      \label{fig:cosx_cond50}
    }
  \end{tabular}
  \caption{Validation of Theorem \ref{theorem:1} and comparison among algorithms with different settings.}
  \label{fig:cosx}
\end{figure*}

To testify the effectiveness of NLISTA and validate our theorem,
we conduct following experiments
where the experiment settings and network training strategies most follow prior works
\cite{yang2016sparse}\cite{chen2018theoretical}\cite{liu2018alista}.

To be more specific, we set $m = 250$ and $n = 500$.
so the dimension of $x^*$, $y$ and $A$ is $500 \times 1$, $250 \times 1$ and $250 \times 500 $.
The elements of $A$ are sampled from Gaussian distribution with variance $\frac{1}{m}$,
and each column of $A$ is normalized to have the unit $\ell_{2}$-norm,
which ensures $A \in \Omega_A$.
The matrix $A$ is fixed in each setting where different algorithms are compared.
The elements of $x^*$ follow the Bernoulli distribution to be zero or non-zero
with the probability being 0.1, 
and the non-zero elements of $x^*$ are sampled from the standard Gaussian distribution.
The nonlinear function is set as $f(x) = 2x + cos(x)$
which is same with \cite{yang2016sparse}\cite{yang2018parallel}.
Therefore, the nonlinear function is continuously differentiable, nonconvex
and the gradient is always nonzero. 
The vector $y$ is generated as (\ref{nonlinear system}) where the noise $\varepsilon$ obeys the Gaussian distribution
with a certain variance according to the signal-to-noise (SNR) ratio which is set infinity as default.
So the setups ensure that there exists a constant $c_x$ , a constant $s$ and a constant $\sigma$
subject to $\forall (x^*,\varepsilon) \in X(c_x,s,\sigma)$.

We randomly synthesize in-stream $x^*$, $\varepsilon$ and $y$ for training and validating.
The training batch size is 64.
The test set $\{ x^* \}$ contains 1000 samples generated as described above,
which is fixed for all tests in our simulations.
All the compared networks have 16 layers and are trained from layer to layer in a progressive way
same with \cite{chen2018theoretical}\cite{liu2018alista}.
For NLISTA, the front eleven layers are fixed when we train the last five layers.
All learnable parameters are not shared among different layers in networks.
The training loss function is $\mathbb{E}\left\|x^{(t)}-x^{*}\right\|_{2}^{2}$
The optimizer is Adam \cite{kingma2014adam}.
The learning rate is first initialized to be 0.001
and will then decrease to 0.0001 and finally decrease to 0.00002,
if the validation loss does not decrease for 4000 iterations. 
To evaluate the recovery performance, we use the normalized mean square error ( NMSE ) in dB:
\begin{equation}
  \label{NMSE}
  \operatorname{NMSE}\left(x^{(t)}, x^{*}\right) = 
  10 \log _{10}\left(\frac{\mathbb{E}\left\|x^{(t)}-x^{*}\right\|_{2}^{2}}{\mathbb{E}\left\|x^{*}\right\|_{2}^{2}}\right),
\end{equation}
where $x^{(t)}$ is the output of the t-th iteration and $x^*$ is the ground truth. 


The baseline algorithms are SpaRSA\cite{yang2016sparse}, FISTA\cite{beck2009fast}, FPCA~\cite{hale2008fixed},
STELA\cite{yang2018parallel} and LISTA\cite{gregor2010learning}.
The regularization parameter for each iterative algorithm is chosen specifically in each experiment.
The detailed description of baseline algorithms and corresponding parameters can be found in the supplementary.
Other learned algorithms such as LAMP\cite{borgerding2017amp}, LISTA-cpss\cite{chen2018theoretical} and ALISTA\cite{liu2018alista} 
are not compared with NLISTA because we find that they are not able to deal with nonlinear cases.

The experiment results under the noiseless condition
are reported in Fig.~\ref{fig:cosx_noiseless},
where NLISTA outperforms other algorithms significantly.
Moreover, the results support Theorem~\ref{theorem:1} that there exists a set of parameters for NLISTA
enabling the upper bound of the recovery error converges to zero at a linear rate.

The experiment results under the noisy condition
are reported in Fig.~\ref{fig:cosx_SNR30},
which demonstrate the robustness of NLISTA to deal with noisy cases
and improvement compared to other algorithms.
Contrasting the results of NLISTA in Fig.~\ref{fig:cosx_SNR30} with Fig.~\ref{fig:cosx_noiseless},
the final recovery error converges exponentially to zero in the noiseless case
and converges to a stationary level related with the noise-level in the noisy case,
which validates the discussion about the influence of the noise after Theorem~\ref{theorem:1}. 

To demonstrate the robustness of NLISTA to deal with ill-conditioned matrices,
we set the condition number of the matrix $A$ equalling to 50.
In Figure \ref{fig:cosx_cond50}, the results show that 
NLISTA still outperforms other algorithms significantly with the ill-conditioned matrices.

In order to explore the influence of nonlinear functions,
we compare $f(x) = 10x+cos(2x)$,$f(x) = 10x+cos(3x)$ and $f(x) = 10x+cos(4x)$,
where the main difference is the supremum of $|\nabla f(x)|$
and all gradients are nonzero for any $x$.
In Table \ref{table: gradient}, the results show that 
the recovery error of NLISTA converges faster with the smaller supremum of $\nabla f(x)$,
which supports the discussion about the upper bound of $|\nabla f(x)|$ after Theorem \ref{theorem:1}.
The law also holds for other algorithms,
which reveals the impact of nonlinear functions on algorithm performance for sparse nonlinear regression problems.
The experiment results of other algorithms are not displayed due to the space limitation
and can be found in the arXiv version of the paper.
The performance of NLISTA is always the best among all algorithms.

\begin{threeparttable}[t]
    \centering
    \caption{Comparison among different gradient supremum.}
    \label{table: gradient}
    \begin{tabular}{p{3cm}<{\centering}p{3cm}<{\centering}
        p{1.5cm}<{\centering}p{1cm}<{\centering}p{1cm}<{\centering}
        p{1cm}<{\centering}p{1cm}<{\centering}p{1.5cm}<{\centering}}
     \toprule
     $f(x)$& $sup(|\nabla f(x)|)$\tnote{1} & SpaRSA & FISTA &FPCA &STELA & LISTA & NLISTA \\
     \midrule
     $10x+cos(2x)$ & 12 & -14.0 & -17.4 & -14.2 & -13.5 & -19.7 & \textbf{-35.7} \\
     $10x+cos(3x)$ & 13 & -13.2 & -16.5 & -13.4 & -12.7 & -16.8 & \textbf{-32.2} \\
     $10x+cos(4x)$ & 14 & -12.4 & -15.3 & -12.5 & -11.8 & -15.7 & \textbf{-28.4} \\
     \bottomrule
    \end{tabular}
    \begin{tablenotes}
      \item[1]$sup(|\nabla f(x)|)$ represents the supremum of $|\nabla f(x)|$.
    \end{tablenotes}
   \end{threeparttable}
 
\section{Conclusion}
\vspace{-0.1em}
In this article, we first unfold the SpaRSA method to solve the sparse nonlinear regression problem,
and we have proposed a new algorithm called NLISTA whose performance is better than existing state-of-art algorithms.
Moreover, we have proved theoretically that there exists a set of parameters enabling NLISTA to converge linearly.
The experiment results support our theorem and analysis and show that such parameters can be learned.
We plan on dealing with the situation that the nonlinear function $f(\cdot)$ is not element-wise,
where the gradient of the nonlinear function $f(\cdot)$ is more complicated.

\newpage
\bibliographystyle{IEEEbib}
\bibliography{reference}

\begin{thebibliography}{10}

\bibitem{blumensath2008iterative}
Thomas Blumensath and Mike~E Davies,
\newblock ``Iterative thresholding for sparse approximations,''
\newblock {\em Journal of Fourier analysis and Applications}, vol. 14, no. 5-6,
  pp. 629--654, 2008.

\bibitem{wright2009sparse}
Stephen~J Wright, Robert~D Nowak, and M{\'a}rio~AT Figueiredo,
\newblock ``Sparse reconstruction by separable approximation,''
\newblock {\em IEEE Transactions on Signal Processing}, vol. 57, no. 7, pp.
  2479--2493, 2009.

\bibitem{beck2009fast}
Amir Beck and Marc Teboulle,
\newblock ``A fast iterative shrinkage-thresholding algorithm for linear
  inverse problems,''
\newblock {\em SIAM journal on imaging sciences}, vol. 2, no. 1, pp. 183--202,
  2009.

\bibitem{hale2008fixed}
Elaine~T Hale, Wotao Yin, and Yin Zhang,
\newblock ``Fixed-point continuation for $\ell_1$-minimization: Methodology and
  convergence,''
\newblock {\em SIAM Journal on Optimization}, vol. 19, no. 3, pp. 1107--1130,
  2008.

\bibitem{yang2018parallel}
Yang Yang, Marius Pesavento, Symeon Chatzinotas, and Bj{\"o}rn Ottersten,
\newblock ``Parallel and hybrid soft-thresholding algorithms with line search
  for sparse nonlinear regression,''
\newblock in {\em 2018 26th European Signal Processing Conference (EUSIPCO)}.
  IEEE, 2018, pp. 1587--1591.

\bibitem{yang2016sparse}
Zhuoran Yang, Zhaoran Wang, Han Liu, Yonina Eldar, and Tong Zhang,
\newblock ``Sparse nonlinear regression: Parameter estimation under
  nonconvexity,''
\newblock in {\em International Conference on Machine Learning}, 2016, pp.
  2472--2481.

\bibitem{gregor2010learning}
Karol Gregor and Yann LeCun,
\newblock ``Learning fast approximations of sparse coding,''
\newblock in {\em Proceedings of the 27th International Conference on
  International Conference on Machine Learning}, 2010, pp. 399--406.

\bibitem{zhang2018ista}
Jian Zhang and Bernard Ghanem,
\newblock ``Ista-net: Interpretable optimization-inspired deep network for
  image compressive sensing,''
\newblock in {\em Proceedings of the IEEE conference on computer vision and
  pattern recognition}, 2018, pp. 1828--1837.

\bibitem{ito2019trainable}
Daisuke Ito, Satoshi Takabe, and Tadashi Wadayama,
\newblock ``Trainable ista for sparse signal recovery,''
\newblock {\em IEEE Transactions on Signal Processing}, vol. 67, no. 12, pp.
  3113--3125, 2019.

\bibitem{ablin2019learning}
Pierre Ablin, Thomas Moreau, Mathurin Massias, and Alexandre Gramfort,
\newblock ``Learning step sizes for unfolded sparse coding,''
\newblock in {\em Advances in Neural Information Processing Systems}, 2019, pp.
  13100--13110.

\bibitem{kim2020element}
Dohyun Kim and Daeyoung Park,
\newblock ``Element-wise adaptive thresholds for learned iterative shrinkage
  thresholding algorithms,''
\newblock {\em IEEE Access}, vol. 8, pp. 45874--45886, 2020.

\bibitem{wusparse}
Kailun Wu, Yiwen Guo, Ziang Li, and Changshui Zhang,
\newblock ``Sparse coding with gated learned ista,''
\newblock in {\em International Conference on Learning Representations}, 2020.

\bibitem{hershey2014deep}
John~R Hershey, Jonathan~Le Roux, and Felix Weninger,
\newblock ``Deep unfolding: Model-based inspiration of novel deep
  architectures,''
\newblock {\em arXiv preprint arXiv:1409.2574}, 2014.

\bibitem{chamon2019sparse}
Luiz~FO Chamon, Yonina~C Eldar, and Alejandro Ribeiro,
\newblock ``Sparse recovery over nonlinear dictionaries,''
\newblock in {\em ICASSP 2019-2019 IEEE International Conference on Acoustics,
  Speech and Signal Processing (ICASSP)}. IEEE, 2019, pp. 4878--4882.

\bibitem{chamon2020functional}
Luiz~FO Chamon, Yonina~C Eldar, and Alejandro Ribeiro,
\newblock ``Functional nonlinear sparse models,''
\newblock {\em IEEE Transactions on Signal Processing}, vol. 68, pp.
  2449--2463, 2020.

\bibitem{chen2018theoretical}
Xiaohan Chen, Jialin Liu, Zhangyang Wang, and Wotao Yin,
\newblock ``Theoretical linear convergence of unfolded ista and its practical
  weights and thresholds,''
\newblock in {\em Advances in Neural Information Processing Systems}, 2018, pp.
  9061--9071.

\bibitem{liu2018alista}
Jialin Liu, Xiaohan Chen, Zhangyang Wang, and Wotao Yin,
\newblock ``{ALISTA}: Analytic weights are as good as learned weights in
  {LISTA},''
\newblock in {\em International Conference on Learning Representations}, 2019.

\bibitem{kingma2014adam}
Diederik~P Kingma and Jimmy Ba,
\newblock ``Adam: A method for stochastic optimization,''
\newblock {\em arXiv preprint arXiv:1412.6980}, 2014.

\bibitem{borgerding2017amp}
Mark Borgerding, Philip Schniter, and Sundeep Rangan,
\newblock ``Amp-inspired deep networks for sparse linear inverse problems,''
\newblock {\em IEEE Transactions on Signal Processing}, vol. 65, no. 16, pp.
  4293--4308, 2017.

\end{thebibliography}

\newpage
\pagenumbering{arabic}
\setcounter{page}{1}
\appendix
\begin{center}
{\Large \centering 
\textbf{Learning Fast Approximations of Sparse Nonlinear Regression (Supplementary Material)}} 
\end{center}

\section{Proof of Lemma \ref{lemma: W}}

\begin{proof}
Since $f(\cdot)$ is a element-wise function, the gradient of $f(\cdot)$ is a diagonal matrix.
And the gradient of $f(\cdot)$  is an invertible matrix because it is nonzero for any $x \in [-c_x,c_x]$.
Let
\begin{equation}
    W \triangleq \frac{1}{\beta^{(t)}} \frac{1}{\gamma^{(t)}} (\nabla f(Ax^{(t)}))^{-1} (\nabla f(\xi^{(t)}))^{-1}  A ,
\end{equation}
then we have
\begin{equation}
    {W_i^{(t)}}^{\mathrm{T}} = \frac{1}{\beta^{(t)}} \frac{1}{\gamma^{(t)}} A_i^{\mathrm{T}} (\nabla f(\xi^{(t)})^{\mathrm{T}})^{-1} (\nabla f(Ax^{(t)})^{\mathrm{T}} )^{-1} .
\end{equation}
Since the gradient of $f(\cdot)$ is a diagonal matrix,
we have
\begin{equation}
    {W_i^{(t)}}^{\mathrm{T}} = \frac{1}{\beta^{(t)}} \frac{1}{\gamma^{(t)}} A_i^{\mathrm{T}} (\nabla f(\xi^{(t)}))^{-1} (\nabla f(Ax^{(t)}) )^{-1} ,
\end{equation}
which means
\begin{equation}
    \beta^{(t)} \gamma^{(t)} {W_i^{(t)}}^{\mathrm{T}} \nabla f(Ax^{(t)}) \nabla f(\xi^{(t)}) A_j = A_i^{\mathrm{T}} A_j.
\end{equation}
Since Assumption \ref{assume:A} holds,
we have
\begin{equation}
    \beta^{(t)} \gamma^{(t)} {W_i^{(t)}}^{\mathrm{T}} \nabla f(Ax^{(t)}) \nabla f(\xi^{(t)}) A_i = 1, i= 1,2,\cdots,n,
\end{equation}
\begin{equation}
    \mathop{\rm max}\limits_{i \neq j} 
    | \beta^{(t)} \gamma^{(t)} {W_i^{(t)}}^{\mathrm{T}} \nabla f(Ax^{(t)}) \nabla f(\xi^{(t)}) A_j | < 1, i,j= 1,2,\cdots,n,
\end{equation}
which means $ W \in \Omega_W^{(t)}$.
Thus $\Omega_W^{(t)}$ is not a empty set.

\end{proof}

\section{Proof of Lemma \ref{lemma: no false positive}}

\begin{proof}
Since $x^{(0)} = 0$, $x_i^{(0)} = 0$ is satisfied for any $i \notin S$.
Fixing $t$, and assuming $x_i^{(t)} = 0$ is satisfied for any $i \notin S$, 
we have
\[
\begin{aligned}
    x_i^{(t+1)} =& \eta(x^{(t)} + \beta^{(t)} \gamma^{(t)} {W_i^{(t)}}^{\mathrm{T}} \nabla f(Ax^{(t)})(y - f(Ax^{(t)})), \theta^{(t)})
    \\
    =& \eta(\beta^{(t)} \gamma^{(t)} {W_i^{(t)}}^{\mathrm{T}} \nabla f(Ax^{(t)})(y - f(Ax^{(t)})), \theta^{(t)})
    \\
    =& \eta(\beta^{(t)} \gamma^{(t)} {W_i^{(t)}}^{\mathrm{T}} \nabla f(Ax^{(t)})(f(Ax^*) - f(Ax^{(t)})) + 
            \beta^{(t)} \gamma^{(t)} {W_i^{(t)}}^{\mathrm{T}} \nabla f(Ax^{(t)}) \varepsilon, \theta^{(t)})
\end{aligned}
\]
Since Lemma \ref{lemma: mean value theorem} holds,
we have
\begin{equation}
    x_i^{(t+1)} = \eta(\beta^{(t)} \gamma^{(t)} {W_i^{(t)}}^{\mathrm{T}} \nabla f(Ax^{(t)}) \nabla f(\xi^{(t)})(Ax^* - Ax^{(t)}) + 
                  \beta^{(t)} \gamma^{(t)} {W_i^{(t)}}^{\mathrm{T}} \nabla f(Ax^{(t)}) \varepsilon, \theta^{(t)}).
\end{equation}
Since
\begin{equation}
    \mu_1^{(t)} = \mathop{\rm max}\limits_{i \neq j} | \beta^{(t)} \gamma^{(t)} {W_i^{(t)}}^{\mathrm{T}} \nabla f(Ax^{(t)}) \nabla f(\xi^{(t)}) A_j | ,
\end{equation}
\begin{equation}
  \mu_2^{(t)} = \mathop{\rm max}\limits_{i} \| \beta^{(t)} \gamma^{(t)} {W_i^{(t)}}^{\mathrm{T}} \nabla f(Ax^{(t)}) \|_1 ,
\end{equation}
and
\begin{equation}
    \theta^{(t)} = \mu_1^{(t)} \|x^* - x^{(t)} \|_1 + \mu_2^{(t)} \sigma,
\end{equation}
we have
\[
\begin{aligned}
    & \quad 
    | \beta^{(t)} \gamma^{(t)} {W_i^{(t)}}^{\mathrm{T}} \nabla f(Ax^{(t)}) \nabla f(\xi^{(t)})(Ax^* - Ax^{(t)}) + 
    \beta^{(t)} \gamma^{(t)} {W_i^{(t)}}^{\mathrm{T}} \nabla f(Ax^{(t)}) \varepsilon |
    \\
    \le & \quad
    | \sum_{j\in S} \beta^{(t)} \gamma^{(t)} {W_i^{(t)}}^{\mathrm{T}} \nabla f(Ax^{(t)}) \nabla f(\xi^{(t)}) A_j (x_j^* - x_j^{(t)}) | + 
    |\beta^{(t)} \gamma^{(t)} {W_i^{(t)}}^{\mathrm{T}} \nabla f(Ax^{(t)}) \varepsilon |
    \\
    \le & \quad
    \sum_{j\in S} | \beta^{(t)} \gamma^{(t)} {W_i^{(t)}}^{\mathrm{T}} \nabla f(Ax^{(t)}) \nabla f(\xi^{(t)}) A_j | |x_j^* - x_j^{(t)}| +
    \| \beta^{(t)} \gamma^{(t)} {W_i^{(t)}}^{\mathrm{T}} \nabla f(Ax^{(t)}) \|_1 \| \varepsilon \|_1
    \\
    \le & \quad
    \mu_1^{(t)} \| x^* - x^{(t)} \|_1 + \mu_2^{(t)} \sigma
    \\
    \le & \quad \theta^{(t)} ,
\end{aligned}
\]
which implies that $x_i^{(t+1)} = 0$ for any $i \notin S$.
By induction, we have
\begin{equation}
    x_i^{(t)} = 0, \quad \forall i \notin S, \quad \forall t \ge 0.
\end{equation}

\end{proof}

\section{Proof of Theorem \ref{theorem:1}}

\begin{proof}

Let $\partial \ell_1(x)$ represent the sub-gradient of $\|x\|_1$ which is a set defined component-wisely:
\begin{equation}
\label{eq:proof_subgradient}
\partial \ell_1(x)_i =\begin{cases}
                        \{\text{sign}(x_i)\}\quad &\text{if $x_i \neq 0$}, \\
                        [-1,1] \quad &\text{if $x_i = 0$}.
                    \end{cases}
\end{equation}
For any $i \in S$, we have
\[
\begin{aligned}
    x_i^{(t+1)} = & \eta(x_i^{(t)} + \beta^{(t)} \gamma^{(t)} {W_i^{(t)}}^{\mathrm{T}} \nabla f(Ax^{(t)}) \nabla f(\xi^{(t)})(Ax^* - Ax^{(t)}) + 
                  \beta^{(t)} \gamma^{(t)} {W_i^{(t)}}^{\mathrm{T}} \nabla f(Ax^{(t)}) \varepsilon, \theta^{(t)})
    \\
    \in & x_i^{(t)} + \beta^{(t)} \gamma^{(t)} {W_i^{(t)}}^{\mathrm{T}} \nabla f(Ax^{(t)}) \nabla f(\xi^{(t)})(Ax^* - Ax^{(t)}) + 
          \beta^{(t)} \gamma^{(t)} {W_i^{(t)}}^{\mathrm{T}} \nabla f(Ax^{(t)}) \varepsilon - 
          \theta^{(t)} \partial \ell_1(x_i^{(t+1)}).
\end{aligned}
\]
We can let $W^{(t)} \in \Omega_W^{(t)}$ because of Lemma \ref{lemma: W}, then we have
\[
\begin{aligned}
      & x_i^{(t)} + \beta^{(t)} \gamma^{(t)} {W_i^{(t)}}^{\mathrm{T}} \nabla f(Ax^{(t)}) \nabla f(\xi^{(t)})(Ax^* - Ax^{(t)})
    \\
    = & x_i^{(t)} + \sum_{j\in S, j \neq i} \beta^{(t)} \gamma^{(t)} {W_i^{(t)}}^{\mathrm{T}} \nabla f(Ax^{(t)}) \nabla f(\xi^{(t)})(A_j x_j^* - A_j x_j^{(t)}) + (x_i^* - x_i^{(t)})
    \\
    = & x_i^* + \sum_{j\in S, j \neq i} \beta^{(t)} \gamma^{(t)} {W_i^{(t)}}^{\mathrm{T}} \nabla f(Ax^{(t)}) \nabla f(\xi^{(t)})A_j (x_j^* - x_j^{(t)}).
\end{aligned}
\]
Then, we have
\begin{equation}
    x_i^{(t+1)} - x_i^* \in   \sum_{j\in S, j \neq i} \beta^{(t)} \gamma^{(t)} {W_i^{(t)}}^{\mathrm{T}} \nabla f(Ax^{(t)}) \nabla f(\xi^{(t)})A_j (x_j^* - x_j^{(t)}) + 
                            \beta^{(t)} \gamma^{(t)} {W_i^{(t)}}^{\mathrm{T}} \nabla f(Ax^{(t)}) \varepsilon - 
                            \theta^{(t)} \partial \ell_1(x_i^{(t+1)}).
\end{equation}
By the definition of $\partial \ell_1(x)$, we have
\begin{equation}
    |x_i^{(t+1)} - x_i^*| \le 
    \sum_{j\in S, j \neq i} | \beta^{(t)} \gamma^{(t)} {W_i^{(t)}}^{\mathrm{T}} \nabla f(Ax^{(t)}) \nabla f(\xi^{(t)})A_j || (x_j^* - x_j^{(t)}) | + 
    |\beta^{(t)} \gamma^{(t)} {W_i^{(t)}}^{\mathrm{T}} \nabla f(Ax^{(t)}) \varepsilon| + \theta^{(t)}
\end{equation}
Since
\begin{equation}
    \mu_1^{(t)} = \mathop{\rm max}\limits_{i \neq j} | \beta^{(t)} \gamma^{(t)} {W_i^{(t)}}^{\mathrm{T}} \nabla f(Ax^{(t)}) \nabla f(\xi^{(t)}) A_j | ,
    i,j= 1,2,\cdots,n,
\end{equation}
and
\begin{equation}
    \mu_2^{(t)} = \mathop{\rm max}\limits_{i} \| \beta^{(t)} \gamma^{(t)} {W_i^{(t)}}^{\mathrm{T}} \nabla f(Ax^{(t)}) \|_1 ,
    i= 1,2,\cdots,n,
\end{equation}
we have
\[
\begin{aligned}
    |x_i^{(t+1)} - x_i^*| \le &
    \mu_1^{(t)} \sum_{j\in S, j \neq i} | (x_j^* - x_j^{(t)}) | + 
    |\beta^{(t)} \gamma^{(t)} {W_i^{(t)}}^{\mathrm{T}} \nabla f(Ax^{(t)}) \varepsilon| + \theta^{(t)}
    \\ \le &
    \mu_1^{(t)} \sum_{j\in S, j \neq i} | (x_j^* - x_j^{(t)}) | + 
    \|\beta^{(t)} \gamma^{(t)} {W_i^{(t)}}^{\mathrm{T}} \nabla f(Ax^{(t)})\|_1 \| \varepsilon \|_1 + \theta^{(t)}
    \\ \le &
    \mu_1^{(t)} \sum_{j\in S, j \neq i} | x_j^* - x_j^{(t)} | + \mu_2^{(t)} \sigma + \theta^{(t)}
\end{aligned}
\]
Let 
\begin{equation}
    \theta^{(t)} = \mu_1^{(t)} \|x^* - x^{(t)} \|_1 + \mu_2^{(t)} \sigma,
\end{equation}
then Lemma \ref{lemma: no false positive} implies
\begin{equation}
    \|x^{(t+1)} - x^* \|_1 = \sum_{i\in S} | x_i^{(t+1)} - x_i^* |.
\end{equation} 
Thus,
\[
\begin{aligned}
    \|x^{(t+1)} - x^*\|_1 \le &
    \sum_{i\in S} (\mu_1^{(t)} \sum_{j\in S, j \neq i} | x_j^* - x_j^{(t)} | + \mu_2^{(t)} \sigma + \theta^{(t)})
    \\ = &
    \mu_1^{(t)}(s-1) \| x^{(t)} - x^*\|_1 + s(\mu_2^{(t)} \sigma + \theta^{(t)})
    \\ = &
    \mu_1^{(t)}(2s-1)\| x^{(t)} - x^*\|_1 + 2s\mu_2^{(t)} \sigma.
\end{aligned}
\]
Let
\begin{equation}
    {\mathop{{\mu}}\limits^{ \sim }}^{(t)}_1 = max(\mu_1^{(0)},\mu_1^{(1)},\cdots,\mu_1^{(t)}),
\end{equation}
and 
\begin{equation}
    {\mathop{{\mu}}\limits^{ \sim }}^{(t)}_2 = max(\mu_2^{(0)},\mu_2^{(1)},\cdots,\mu_2^{(t)}),
\end{equation}
then we have
\begin{equation}
    \|x^{(t+1)} - x^*\|_1 \le
    ({\mathop{{\mu}}\limits^{ \sim }}^{(t)}_1 (2s-1))^{t+1} \| x^0 - x^*\|_1 +
    2s {\mathop{{\mu}}\limits^{ \sim }}^{(t)}_2 \sigma
    \sum_{i = 0}^{t} ({\mathop{{\mu}}\limits^{ \sim }}^{(t)}_1 (2s-1))^{i}.
\end{equation}
Since $x^{(0)} = 0$, and $x^* \in \Omega_x(c_x,s)$,
we have
\begin{equation}
    \|x^{(t+1)} - x^*\|_1 \le
    ({\mathop{{\mu}}\limits^{ \sim }}^{(t)}_1 (2s-1))^{t+1} s c_x+
    2s {\mathop{{\mu}}\limits^{ \sim }}^{(t)}_2 \sigma
    \sum_{i = 0}^{t} ({\mathop{{\mu}}\limits^{ \sim }}^{(t)}_1 (2s-1))^{i}.
\end{equation}
Since $W^{(t)} \in \Omega_W^{(t)}$, we have $\mu_1^{(t)} < 1$ for any $t \ge 0$.
Thus ${\mathop{{\mu}}\limits^{ \sim }}^{(t)}_1 < 1$.
With $q = {\mathop{{\mu}}\limits^{ \sim }}^{(t)}_1(2s-1)$,
and $c_{\varepsilon} =  {\mathop{{\mu}}\limits^{ \sim }}^{(t)}_2
\sum_{i = 0}^{t} ({\mathop{{\mu}}\limits^{ \sim }}^{(t)}_1 (2s-1))^{i}$,
we have 
\begin{equation}
    \|x^{(t+1)} - x^*\|_2 \le
    \|x^{(t+1)} - x^*\|_1 \le
    q^{t+1} s c_x + 2 c_{\varepsilon} s  \sigma,
\end{equation}
where $q \in (0,1)$ when $s \in [1, \frac{1}{2}(( {\mathop{{\mu}}\limits^{ \sim }}^{(t)}_1 )^{-1} + 1))$.

\end{proof}

\section{The Details of Baseline Algorithms}

The SpaRSA (Sparse Reconstruction by Separable Approximation) method with the line search procedure to choose $\alpha^{(t)}$ 
is given in Algorithm \ref{algo:SpaRSA},
which avoids calculating the eigenvalues of $\nabla^2 L(x^{(t)})$.
We take $\eta=2$, $\xi=10^{-5}$ and $M=0$ in our experiments,
whose effect on the experimental results is not significant.
We take $\lambda = 0.5$ for $f(x)=2x+cos(x)$, 
$\lambda = 11$ for $f(x)=10x+cos(2x)$,
$\lambda = 12$ for $f(x)=10x+cos(3x)$,
and $\lambda = 12$ for $f(x)=10x+cos(4x)$,
which are almost the optimal choices.

\begin{algorithm2e}[htb]
    \SetKwInOut{input}{Input}
    \SetKwInOut{initial}{Initialization}
    \input{dictionary matrix $A$,
        vector $y$,
        regularization parameter $\lambda > 0$,
        nonlinear function $f(x)$,
        error function $L(x):= \frac{1}{2}\|y-f(Ax)\|_{2}^{2}$,
        loss function $\phi(x) := L(x)+\lambda\|x\|_{1}$,
        update factor $\eta > 1$,
        constant $\xi>0$, constant $M \ge 0$,
        and maximum iteration $T>0$
        }
    \initial{set $x^{(0)} \leftarrow \textbf{0}$}
    \For{$t = 0,1,\cdots,T-1$}
        {
        Choose $\alpha^{(t)}$ according to Algorithm \ref{algo:BB}
        \\
        $ x^{(t+1)} \leftarrow \eta(x^{(t)} - \frac{1}{\alpha^{(t)}} \nabla L(x^{(t)}), \frac{\lambda}{\alpha^{(t)}} ) $
        \\
        \While{
            $\phi(x^{(t+1)}) > 
            \mathop{\rm max}\limits_{max(t-M,0) \le j \le t}
            \{ \phi(x^{(j)}) - \xi \frac{\alpha^{(t)}}{2} \|x^{(t+1)} - x^{(t)} \|_2^2 \}$
        }
        {
            $ \alpha^{(t)} \leftarrow \eta \alpha^{(t)} $
            \\
            $ x^{(t+1)} \leftarrow \eta(x^{(t)} - \frac{1}{\alpha^{(t)}} \nabla L(x^{(t)}), \frac{\lambda}{\alpha^{(t)}} ) $
        }
        } 
    \KwOut{$x^{(T)}$}
    \caption{The SpaRSA (Sparse Reconstruction by Separable Approximation) method}
    \label{algo:SpaRSA}
\end{algorithm2e}
\vspace{-1em}
\begin{algorithm2e}[htb]
    \SetKwInOut{input}{Input}
    \SetKwInOut{initial}{Initialization}
    \input{iteration counter $t$, $x^{(t)}$,$x^{(t-1)}$ and error function $L(x)$} 
    \initial{Let $\delta^{(t)} \leftarrow x^{(t)} - x^{(t-1)}$ 
            and $g^{(t)} \leftarrow \nabla L(x^{(t)}) - \nabla L(x^{(t-1)})$
            }
        \uIf{$t = 0$}{ \KwOut{$\alpha^{(t)} \leftarrow 1$}
        }\Else { \KwOut{
            $\alpha^{(t)} \leftarrow \frac{(\delta^{(t)})^{\mathrm{T}} g^{(t)}}{(g^{(t)})^{\mathrm{T}} g^{(t)}} $
            or
            $\alpha^{(t)} \leftarrow \frac{(g^{(t)})^{\mathrm{T}} g^{(t)}}{(\delta^{(t)})^{\mathrm{T}} g^{(t)}} $
            }
        }
    \caption{The Barzilai-Borwein (BB) spectral approach for choosing $\alpha^{(t)}$}
    \label{algo:BB}
\end{algorithm2e}

\newpage
\begin{algorithm2e}[t]
    \SetKwInOut{input}{Input}
    \SetKwInOut{initial}{Initialization}
    \input{dictionary matrix $A$,
        vector $y$,
        regularization parameter $\lambda > 0$,
        nonlinear function $f(x)$,
        error function $L(x):= \frac{1}{2}\|y-f(Ax)\|_{2}^{2}$,
        loss function $\phi(x) := L(x)+\lambda\|x\|_{1}$,
        update factor $\eta > 1$,
        constant $\xi>0$, constant $M \ge 0$,
        and maximum iteration $T>0$
        }
    \initial{set $x^{(0)} \leftarrow \textbf{0}$, $z^{(1)} \leftarrow x^{(0)}$, $k^{(0)} \leftarrow 1$}
    \For{$t = 0,1,\cdots,T-1$}
        {
        Choose $\alpha^{(t)}$ according to Algorithm \ref{algo:BB}
        \\
        $ x^{(t+1)} \leftarrow \eta(z^{(t+1)} - \frac{1}{\alpha^{(t)}} \nabla L(z^{(t+1)}), \frac{\lambda}{\alpha^{(t)}} ) $
        \\
        \While{
            $\phi(x^{(t+1)}) > 
            \mathop{\rm max}\limits_{max(t-M,0) \le j \le t}
            \{ \phi(x^{(j)}) - \xi \frac{\alpha^{(t)}}{2} \|x^{(t+1)} - x^{(t)} \|_2^2 \}$
        }
        {
            $ \alpha^{(t)} \leftarrow \eta \alpha^{(t)} $
            \\
            $ x^{(t+1)} \leftarrow \eta(x^{(t)} - \frac{1}{\alpha^{(t)}} \nabla L(x^{(t)}), \frac{\lambda}{\alpha^{(t)}} ) $
        }
        $ k^{(t+1)} \leftarrow \frac{1 + \sqrt{1+4(k^{(t)})^2}}{2} $
        \\
        $ z^{(t+2)} \leftarrow x^{(t+1)} + \frac{k^{(t)}-1}{k^{(t+1)}}(x^{(t+1)} - x^{(t)})  $
        \\
        } 
    \KwOut{$x^{(T)}$}
    \caption{The Fast Iterative Soft Thresholding Algorithm (FISTA)}
    \label{algo:FISTA}
\end{algorithm2e}
The fast iterative soft thresholding algorithm (FISTA) with the line search procedure to choose $\alpha^{(t)}$ 
is given in Algorithm \ref{algo:FISTA}.
We take $\eta=2$, $\xi=10^{-5}$ and $M=0$ in our experiments,
whose effect on the experimental results is not significant.
We take $\lambda = 0.4$ for $f(x)=2x+cos(x)$, which is almost the optimal choice.

The fixed point continuation algorithm (FPCA) with the line search procedure to choose $\alpha^{(t)}$ 
is given in Algorithm \ref{algo:FPCA}.
We take $\eta=2$, $\xi=10^{-5}$ and $M=0$ in our experiments,
whose effect on the experimental results is not significant.
We take $\lambda = 0.5$ for $f(x)=2x+cos(x)$, 
$\lambda = 8$ for $f(x)=10x+cos(2x)$,
$\lambda = 9$ for $f(x)=10x+cos(3x)$,
and $\lambda = 10$ for $f(x)=10x+cos(4x)$,
which are almost the optimal choices.

The iterative soft thresholding with line search algorithm (STELA) is given in Algorithm \ref{algo:STELA}.
We take $\eta=2$, $\xi=10^{-5}$ and $M=0$ in our experiments,
whose effect on the experimental results is not significant.
We take $\lambda = 0.5$ for $f(x)=2x+cos(x)$,
$\lambda = 11$ for $f(x)=10x+cos(2x)$,
$\lambda = 13$ for $f(x)=10x+cos(3x)$,
and $\lambda = 14$ for $f(x)=10x+cos(4x)$,
which are almost the optimal choices.

\begin{algorithm2e}[h]
    \SetKwInOut{input}{Input}
    \SetKwInOut{initial}{Initialization}
    \input{dictionary matrix $A$,
        vector $y$,
        regularization parameter $\lambda > 0$,
        nonlinear function $f(x)$,
        error function $L(x):= \frac{1}{2}\|y-f(Ax)\|_{2}^{2}$,
        loss function $\phi(x) := L(x)+\lambda\|x\|_{1}$,
        update factor $\eta > 1$,
        constant $\xi>0$, constant $M \ge 0$,constant $\gamma > 0$,
        and maximum iteration $T>0$
        }
    \initial{set $x^{(0)} \leftarrow \textbf{0}$}
    \For{$t = 0,1,\cdots,T-1$}
        {
        Choose $\alpha^{(t)}$ according to Algorithm \ref{algo:BB}
        \\
        $ x^{(t+1)} \leftarrow \eta(x^{(t)} - \frac{1}{\alpha^{(t)}} \nabla L(x^{(t)}), \frac{(\lambda}{\alpha^{(t)}} ) $
        \\
        \While{
            $\phi(x^{(t+1)}) > 
            \mathop{\rm max}\limits_{max(t-M,0) \le j \le t}
            \{ \phi(x^{(j)}) - \xi \frac{\alpha^{(t)}}{2} \|x^{(t+1)} - x^{(t)} \|_2^2 \}$
        }
        {
            $ \alpha^{(t)} \leftarrow \eta \alpha^{(t)} $
            \\
            $ x^{(t+1)} \leftarrow \eta(x^{(t)} - \frac{1}{\alpha^{(t)}} \nabla L(x^{(t)}), \frac{(\lambda}{\alpha^{(t)}} ) $
        }
        \If{ $ \|x^{(t+1)} - x^{(t)} \|_2 < \gamma $}{
            $\lambda \leftarrow 0.5 \lambda$
            \\
            $\gamma \leftarrow 0.5 \gamma$
        }
        } 
        
    \KwOut{$x^{(T)}$}
    \caption{The Fixed Point Continuation Algorithm (FPCA)}
    \label{algo:FPCA}
\end{algorithm2e}

\begin{algorithm2e}[h]
    \SetKwInOut{input}{Input}
    \SetKwInOut{initial}{Initialization}
    \input{dictionary matrix $A$,
        vector $y$,
        regularization parameter $\lambda > 0$,
        nonlinear function $f(x)$,
        error function $L(x):= \frac{1}{2}\|y-f(Ax)\|_{2}^{2}$,
        loss function $\phi(x) := L(x)+\lambda\|x\|_{1}$,
        update factor $\eta > 1$,
        constant $\xi>0$, constant $\beta \in (0,1)$ ,
        and maximum iteration $T>0$
        }
    \initial{set $x^{(0)} \leftarrow \textbf{0}$, $\gamma^{(0)} \leftarrow 1$}
    \For{$t = 0,1,\cdots,T-1$}
        {
        Choose $\alpha^{(t)}$ according to Algorithm \ref{algo:BB}
        \\
        $ x_{dir} \leftarrow \eta(x^{(t)} - \frac{1}{\alpha^{(t)}} \nabla L(x^{(t)}), \frac{\lambda}{\alpha^{(t)}} ) $
        \\
        \While{
            $L(x^{(t)} + \gamma^{(t)} (x_{dir} - x^{(t)})) + \lambda((1-\gamma^{(t)} ) \|x^{(t)}\|_{1} + \gamma^{(t)} \|x_{dir}\|_{1}  )
                > 
              \phi(x^{(t)}) + \xi \gamma^{(t)} ( {\nabla L(x^{(t)})}^{\mathrm{T}}(x_{dir} - x^{(t)}) ) + \lambda (\|x_{dir}\|_{1} - \|x^{(t)}\|_{1}) $
        }
        {
            $\gamma^{(t)} \leftarrow  \beta \gamma^{(t)}$
        }
        $ x^{(t+1)} = x^{(t)} + \gamma^{(t)} (x_{dir} - x^{(t)}) $
        } 
    \KwOut{$x^{(T)}$}
    \caption{The iterative Soft ThrEsholding with Line search Algorithm (STELA)}
    \label{algo:STELA}
\end{algorithm2e}

\end{document}